\documentclass{article}

\usepackage{graphicx}
\usepackage{amsmath}
\usepackage{amsfonts}
\usepackage[final]{gpsmdms_2022}
\usepackage{booktabs}
\usepackage{caption}
\usepackage{subcaption}
\usepackage{multirow}
\usepackage{blindtext}
\usepackage[capitalize]{cleveref}

\usepackage[usenames,dvipsnames]{xcolor}
\definecolor{cyan}{cmyk}{1,0,0,0.4}
\definecolor{magenta}{cmyk}{0,1,0,0.15}
\definecolor{yellow}{cmyk}{0,0,1,0}
\definecolor{black}{cmyk}{0,0,0,1}
\definecolor{orange}{cmyk}{0,0.5,1,0}
\definecolor{red}{cmyk}{0.00,0.92,0.92,0.06}
\definecolor{blue}{cmyk}{0.92,0.92,0.00,0.06}
\definecolor{green}{cmyk}{0.88,0.00,0.78,0.03}

\newcommand{\vf}{\mathbf{f}}
\newcommand{\vl}{\mathbf{\ell}}
\newcommand{\vx}{\mathbf{x}}
\newcommand{\vX}{\mathbf{X}}
\newcommand{\vy}{\mathbf{y}}
\newcommand{\vz}{\mathbf{z}}

\newcommand{\vsigma}{\mathbf{\sigma}}
\newcommand{\vtheta}{\mathbf{\theta}}
\newcommand{\vphi}{\mathbf{\phi}}
\newcommand{\vomega}{\mathbf{\omega}}

\newcommand{\inducingX}{\overline{\vX}}
\newcommand{\inducingY}{\overline{\vz}}

\newcommand{\kernelfn}{\mathcal{K}}
\newcommand{\real}{\mathbb{R}}
\newcommand{\eat}[1]{}
\newcommand{\var}{\text{var}}
\newcommand{\cov}{\text{cov}}
\newcommand{\gauss}{\mathcal{N}}

\title{Uncertainty Disentanglement with Non-stationary Heteroscedastic Gaussian Processes for Active Learning}

\begin{document}

\author{
  Zeel B Patel\\
  IIT Gandhinagar, India
  \and
  Nipun Batra\\
  IIT Gandhinagar, India
  \and
  Kevin Murphy\\
  Google, USA
}

\maketitle

\begin{abstract}
    Gaussian processes are Bayesian non-parametric models used in many areas. In this work, we propose a Non-stationary Heteroscedastic Gaussian process model which can be learned with gradient-based techniques. We demonstrate the interpretability of the proposed model by separating the overall uncertainty into aleatoric (irreducible) and epistemic (model) uncertainty. We illustrate the usability of derived epistemic uncertainty on active learning problems. We demonstrate the efficacy of our model with various ablations on multiple datasets.
\end{abstract}

\section{Introduction}
Gaussian processes (GPs) are Bayesian non-parametric models useful for many real-world regression and classification problems.
The key object required to define a GP is the kernel function
$\kernelfn(\vx, \vx')$, which measures the similarity of the input points.
A common choice is the 
% \nb{nitpick: we have not expanded RBF earlier. Given the community, this should not really be an issue.}
RBF kernel $\kernelfn(\vx, \vx'; \vtheta) = \sigma \exp\left(-\frac{||\vx- \vx'||^2}{2\ell^2}\right)$ where $\ell$ is the length scale, and $\sigma$ is the signal variance.
In regression problems, we also often have observation noise
with variance $\omega^2$.
These three hyper-parameters, $\theta=(\ell, \sigma, \omega)$,
are often learned by optimizing the negative log marginal likelihood.

% \nb{it may be useful to quickly define stationarity and homoskedasticity}
However, this model uses a stationary kernel (depended only on the distance between locations) and homoskedastic noise (constant noise variance ($\omega^2$)),
and these assumptions might not hold in real-life applications such as environment modeling~\cite{sampson1992nonparametric,Patel2022AQ}.
In particular, non-stationary kernels are necessary if the similarity of two inputs may depend on their location in the input space.
Similarly, heteroskedastic noise may be necessary if the quality of the measurements may vary across space.

In this short paper, we provide a computationally efficient
way to create GPs with non-stationary kernels, and heteroskedastic noise,
by  using a Gibbs kernel \cite{gibbs1998bayesian}.
% \nb{what is this?! perhaps better to say that the Gibbs kernel can be considered the generalization of the RBF kernel where..}
Gibbs kernel can be considered the generalization of the RBF kernel where the hyper-parameters
are input-dependent, i.e., 
$\theta(\vx) = (\ell(\vx), \sigma(\vx), \omega(\vx))$.
These three hyper-parameter
functions are themselves represented by a
% \nb{wrong quotation marks used}
``latent" GP, as in \cite{Tolvanen2014,heinonen2016non}.
In contrast to prior work, which uses EP or HMC,
we use inducing point approximations to 
% \nb{as we mention speed here we may have to do a timing comparison?!}
speed up the computation of this
latent GP (which is needed to evaluate the kernel).
In addition,
we show how modeling variation in all three hyper-parameters allows us to distinguish locations where the latent function value is uncertain (epistemic uncertainty), as opposed to locations where the observation noise is high (aleatoric uncertainty).
% \nb{it may be useful to introduce the keywords: epistemic and aleatoric here..}. 
This distinction is crucial for problems such as active learning and efficient sensor placement.
% \nb{placement?}
(c.f. \cite{Krause08}).

\section{Methods}

\subsection{Non-stationary Heteroscedastic Gaussian processes}
Given observations $\vy \in \real^N$ at inputs $X \in \real^{N \times D}$, we assume the following model:
\begin{align}
    y(\vx) &= f(\vx)+\varepsilon(\vx), 
    \quad \varepsilon(\vx) \sim \gauss\left(0, \omega(\vx)^2\right) \\
    f(\vx) &\sim GP(0, \kernelfn_f(\vx, \vx'))
\end{align}
where the kernel function is $\kernelfn(\vx, \vx') = \cov(f(\vx), f(\vx'))$ and $\varepsilon(\vx)$ is zero mean noise. We use the following non-stationary kernel function~\cite{gibbs1998bayesian}:
\begin{align}
\kernelfn_f(\vx,\vx')
=  \sigma(\vx) \sigma(\vx')
\sqrt{ \frac{2 \ell(\vx) \ell(\vx')}{\ell(\vx)^2 + \ell(\vx')^2}}
\exp\left(-\frac{||\vx-\vx'||^2}{\ell(\vx)^2 + \ell(\vx')^2} \right)
\label{eqn:gibbsKernel}
\end{align}

We assume all hyperparameters of the model,
$\vtheta(\vx) = (\ell(\vx), \sigma(\vx), \omega(\vx))$,
may be input dependent, to allow for non-stationarity and heteroscedasticity (Previous work~\cite{heinonen2016non} has shown that such a kernel is a valid positive semi-definite kernel). We assume these  ``hyper-functions'' $h(\vx)$
(where $h(\cdot)$ represents either
$\ell(\cdot)$, $\sigma(\cdot)$ or $\omega(\cdot)$)
are smooth and model them by
a latent GP on the log scale:
\begin{align}
\tilde{h}(\vx) \sim GP(\mu_h, \kernelfn_h(\vx, \vx'; \vphi_h))
\end{align}
where $\tilde{h}(\cdot) = \log h(\cdot)$.
These latent GPs are characterized by
a constant mean $\mu_h$
and RBF kernels
with parameters $\vphi_h=(\ell_h, \sigma_h)$.
(We assume noise-free latent functions,
so $\omega_h=0$.)\footnote{
In practice, we use a reasonably small value (jitter) for numerical stability.
% In practice we use $\sigma_h=10^{-4}$ for numerical stability.
% \kevin{Is this correct?} Yes
}

% \nb{do we finally use Sparse GPs with the same inducing locations as the inducing locations for inner GPs?}
To make learning the latent GPs efficient, we will use a set of $M$
(shared) inducing points $\inducingX \in \real^{M \times D}$,
which we treat as additional hyper-parameters.
Let $\inducingY_h = \tilde{h}(\inducingX)$ be the (log) outputs
at these locations for hyper-parameter $h$.
Then we can infer the expected hyper-parameter value at any other location 
$\vx$ using
the usual GP prediction formula for the mean:
\begin{align}
    \tilde{h}(\vx) = \kernelfn_h(\vx, \inducingX)
      \kernelfn_h(\inducingX, \inducingX)^{-1}
      \inducingY_h
\end{align}
Then we can compute $\kernelfn_f(\vx, \vx')$ at any pair of inputs
using \cref{eqn:gibbsKernel},
where $h(\vx) = e^{\tilde{h}(\vx)}$.

\subsection{Learning the hyper-parameters}

In total, we can have up to $2MD+ 2M + 9$ parameters: $MD$-dimensional inducing inputs $\inducingX$, $MD$ sized $\ell(\inducingX)$ (ARD), $M$ sized each $\{\sigma(\inducingX), \omega(\inducingX)$\} and $9$ latent GP hyper-parameters ($\{\mu_h, \vphi_h$\}).
\begin{align}
\vphi = 
% (
% \ell_{\ell}, \sigma_{\ell}, \mu_{\ell},
% \ell_{\sigma}, \sigma_{\sigma}, \mu_{\sigma},
% \ell_{\omega}, \sigma_{\omega}, \mu_{\omega})
(\mu_h, \vphi_h, \inducingY_h)
\end{align}
% Dr. Murphy: (In practice we fix $\mu_h=0$ for all 3 functions.)
% Zeel: No, we are learning mean as well.

Let $\vphi^X = (\inducingX, \vphi)$ represent all the 
model parameters.
We can compute a MAP-type II estimate of these parameters
by minimizing the following objective
using gradient descent:
\begin{align}
    \hat{\vphi}^X = \arg \min_{\vphi^X} -[\log p(\vy|\vX, \vphi, \inducingX)
    + \log p(\vphi)]
\end{align}
where $p(\vy|\vX, \vphi)$ is the marginal likelihood of the
main GP (integrating out
% \nb{also integrating out $\bar{X}$?}
$\vf=[f(\vx_n)]_{n=1}^N$),
and where $p(\vphi)$ is the prior.
To ensure smoothly varying latent GPs with a large length scale
and low variance,
we use a Gamma(5, 1) prior for $\ell$
and a Gamma(0.5, 1) prior for $\sigma$.
% \nb{prior for $\omega$} Zeel: \omega is fixed for latent GP kernels.
% We use a standard normal prior for $\inducingX$,
% after standardizing the inputs.
% \kevin{Is this correct?}
To ease the optimization process, we use a non-centered
parameterization of $\inducingY_h$ by learning a vector $\boldsymbol{\gamma}_h$ with independent values (prior for $\boldsymbol{\gamma}_h$ is $\mathcal{N}(0, 1)$) and we then deterministically compute $\inducingY_h = L\boldsymbol{\gamma}_h$, where, $L$ is Cholesky decomposition of $\kernelfn_h(\inducingX, \inducingX)$.
% \kevin{Give details}
We initialize $\vphi$ by sampling from
their respective priors,
and initialize $\inducingX$ by selecting $M$ points from the dataset.

\eat{
We assume the trend of these functions to be smooth over the input space and thus model them with a latent GP. However, learning hyperparameters at all locations can be expensive and thus we learn them at a few locations $\bar{x}$ (inducing points) and infer the values at all other locations with GP predictive mean estimate,

\begin{align}
&\log (h(\overline{x})) \sim GP_h \left(\mu_h, K_h\left(\overline{x}, \overline{x}^{\prime}\right)\right)\\
&h(x) = \exp\left(K_h(x, \overline{X})K^{-1}(\overline{X}, \overline{X})\log(h(\overline{x}))\right)
\end{align}
Where $h(\cdot)$ represents all $\ell(\cdot), \sigma(\cdot)$ and $\omega(\cdot)$. We use $\exp$ transform to ensure the positivity of the hyperparameters. Learnable parameters of this model are $\{\overline{\ell}_h, \overline{\sigma}_h, \overline{\omega}_h, \overline{X}\}$ where $\overline{\ell}_h, \overline{\sigma}_h, \overline{\omega}_h$ are the hyperparameters of the latent GP and $\overline{X}$ are the inducing locations. We consider the negative log marginal likelihood of the GP as the loss function and optimize it with gradient-based optimization techniques. To regularize the optimization better, we assume priors over the latent GP hyperparameters. We discuss more details on that in the evaluation section.
}

\subsection{Active learning}
\label{sec:AL}

Active learning (see e.g., \cite{settles2009active}) 
uses some measure of uncertainty to decide which points to label
so as to learn the underlying function as quickly as possible.
This can also be useful for tasks such as deciding where to place 
air quality (AQ) sensors
(see e.g., \cite{Patel2022AQ}),
where the goal is to infer the underlying AQ values at all spatial locations
based on sensors of varying quality.

Often the GP predictive variance is used as the acquisition function.
The overall predictive variance,
 $\var(y(\vx))$,
 is equal to the model or epistemic uncertainty,
 $\var(f(\vx))=\sigma(\vx)$,
 plus the observation noise, $\varepsilon(\vx) \sim \gauss(0,\omega(\vx)^2)$,
 which is known as aleatoric uncertainty
 \cite{Hullermeier2021}.
 Aleatoric uncertainty is irreducible,
 and thus querying data-points where it is high is not likely to improve the model
 (c.f., \cite{Osband2016dropout}). Thus, we use the epistemic uncertainty $\var(f(\vx))$ for active learning. (We note that separating epistemic and aleatoric uncertainty is harder with other kinds of probabilistic model, such as Bayesian neural networks,
 c.f., \cite{Valdenegro-Toro2022,Depeweg2018}).

\eat{
One potential application is sensor location recommendation in a hybrid air quality network where model is learned on data from high accuracy sensors and low-cost low-accuracy sensors. In such cases, it is not useful for an active learning strategy to recommend locations with high model uncertainty, because that uncertainty might be irreducible due to low-accuracy sensors. Instead, if we can use the model (epistemic) uncertainty alone, we may be able to recommend better locations. 
}

\section{Experiments}

In this section, we experimentally compare our method to various baselines on various datasets.

\begin{figure}[h]
    \centering
\includegraphics[width=0.7\textwidth]{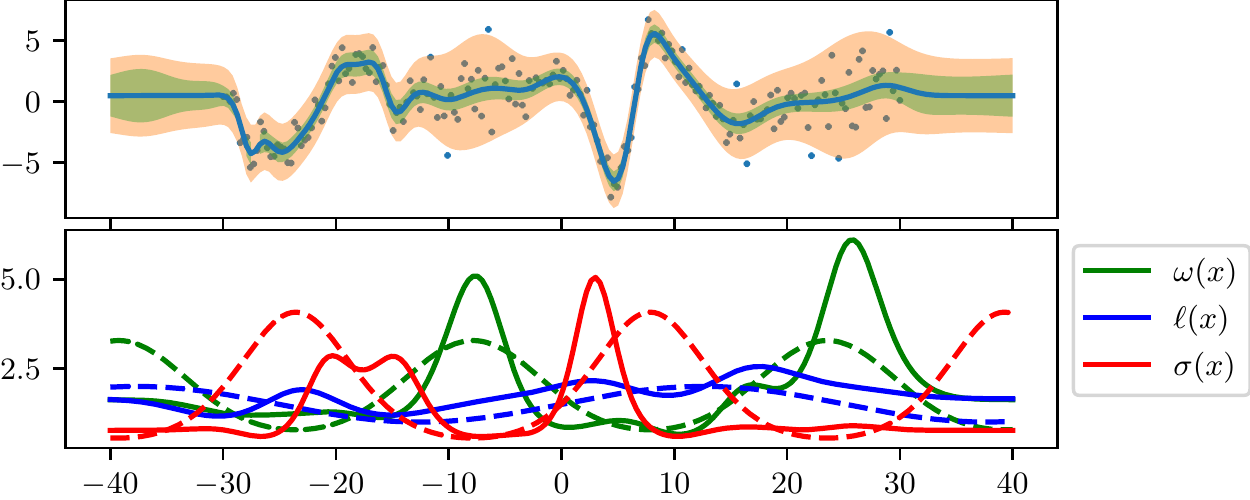}
    \caption{Our model fit on SYNTH-1D data (top) and recovered hyperparameter trends (bottom). Dotted lines show the true functions and continuous lines show the learned functions.}
    \label{fig:recover}
\end{figure}

\paragraph{Regression}
In \cref{tab:results}, we compare various ablations of our method on various small regression tasks. In particular, we consider making one or more of the hyper-parameters corresponding to length scale ($\ell$), variance ($\sigma$), and observation noise ($\omega$) either input-dependent (using a latent GP) or fixed constants to be learned The metrics are averaged over five-fold cross-validation using the Adam optimizer (rate=0.05, epochs=1000). We use the following datasets for experiments: 1) Motorcycle helmet: It is a real-life non-stationary, heteroscedastic dataset used in related studies~\cite{heinonen2016non}; 2) NONSTAT-2D: It is a non-stationary 2D dataset simulated by input-depended lengthscale~\cite{plagemann2008nonstationary}. We add linearly increasing heteroscedastic noise to NONSTAT-2D to make it heteroscedastic; and 3) Jump1D: This is 1D dataset taken from \cite{heinonen2016non} (mentioned as J in \cite{heinonen2016non}), which is highly discontinuous from the center. In general, we get better results by making all these parameters input-dependent. We generate a dataset SYNTH-1D from a generative process of a GP where all three hyperparameters ($\ell(x), \sigma(x), \omega(x)$) are input-depended. We use 200 equally spaced inputs in range (-30, 30) with the following deterministic trends: $\ell(x) = 0.5\sin(x / 8) + 1.5, \sigma(x)=1.5\exp(\sin(0.2x)), \omega(x) = 2.5\log(1+\exp(\sin(-0.2x)))$. We show in \cref{fig:recover} that we are able recover the trends of underlying hyperparameter functions. We illustrate the fit of $(\ell, \sigma, \omega)$-GP over a few more datasets in Appendix Section \ref{sec:appendix} \cref{fig:mcycle}.
% We will release the code at camera-ready.

\renewcommand{\tabcolsep}{2pt}
\begin{table}[ht]
% \addtolength{\tabcolsep}{-2pt}
% \tiny
\fontsize{7}{11}\selectfont
\begin{center}
\begin{tabular}{lrrrrrr}
\toprule
{} & \multicolumn{2}{c}{Jump} & \multicolumn{2}{c}{Motorcycle} & \multicolumn{2}{c}{NONSTAT-2D}\\
Model &              NLPD &             RMSE &             NLPD &             RMSE &               NLPD &             RMSE \\
\midrule
Stationary Homoskedastic GP                  &           4.98 &           0.26 &          11.96 &           0.44 &           -50.72 &           0.09 \\
$(\ell)$-GP               &           5.01 &           0.26 &          11.92 &           0.44 &           -65.13 &           0.06 \\
$(\omega)$-GP             &           3.82 &           0.22 &           5.21 &           0.44 &           -50.81 &           0.09 \\
$(\sigma)$-GP             &          0.92 &           0.30 &          11.56 &           0.44 &           -56.66 &           0.07 \\
$(\ell,\omega)$-GP        &        5.01  &           0.26 &           5.68 &           0.45 &           -65.31 &           0.06 \\
$(\ell,\sigma)$-GP        &          -2.18 &  0.22 &          11.54 &  \textbf{0.44} &           -49.28 &           0.07 \\
$(\sigma,\omega)$-GP      &          0.92 &           0.22 &           4.21 &           0.46 &           -54.35 &           0.10 \\
$(\ell,\sigma,\omega)$-GP &          \textbf{-2.20} &           \textbf{0.22} &  \textbf{4.09} &           0.45 &  \textbf{-73.74} &  \textbf{0.05} \\
\bottomrule
% \toprule
%  & \multicolumn{3}{c}{Motorcycle} & \multicolumn{3}{c}{NONSTAT-2D} & \multicolumn{3}{c}{SYNTH-1D} \\
% Method & RMSE & NLPD & MSLL & RMSE & NLPD & MSLL & RMSE & NLPD & MSLL\\ 
% \midrule
% GP & \textbf{0.44} & 12.22 & 11.95 & 0.12 & -36.32 & -29.39 & 1.23 & 97.09 & 97.81\\
% $(\omega)$-GP & \textbf{0.44} & 5.37 & 5.22 & 0.12 & -36.31 & -32.8 & 1.22 & 88.92 & 90.4 \\
% $(\ell, \omega)$-GP & 0.45 & 6.21 & 6.09 & 0.13 & -48.31 & -39.02 & 1.20 & 88.28 & 89.42\\
% $(\sigma, \omega)$-GP & 0.46 & 4.61 & 4.23 & \textbf{0.11} & -46.26 & -37.69 & 1.22 & 88.60 & 89.82\\
% $(\ell, \sigma, \omega)$-GP & 0.45 & \textbf{4.35} & \textbf{4.06} & \textbf{0.11} & \textbf{-49.15} & \textbf{-39.49} & 1.19 & 88.14 & 89.13\\
% \bottomrule
\end{tabular}
\end{center}
\caption{Quantitative results on various datasets.
Metrics are as follows (lower is better):
Negative Log Predictive Density (NLPD), Root Mean Squared Error (RMSE).
The rows represent different methods in which we either
used fixed or input-dependent length scale $\ell$,
signal variance $\sigma$
and observation noise $\omega$. N-NONSTAT-2D, and other datasets are from \cite{heinonen2016non}. $(\ell,\sigma,\omega)$-GP is the best or the second best across all datasets and metrics.
% \nb{remove accr. and mention full names. mention best/second-best for flexible model}
% \nb{adding other datasets from reference..we want the diagonal in the table to be bolded ideally..also use the convention from the reference paper (l-GP), (l, w GP, etc.)}
% \kevin{Is this correct?} Zeel: Yes
}
\label{tab:results}
\end{table}

\begin{figure}[h]
\centering
\begin{subfigure}{0.4\textwidth}
\centering
\includegraphics[width=\linewidth]{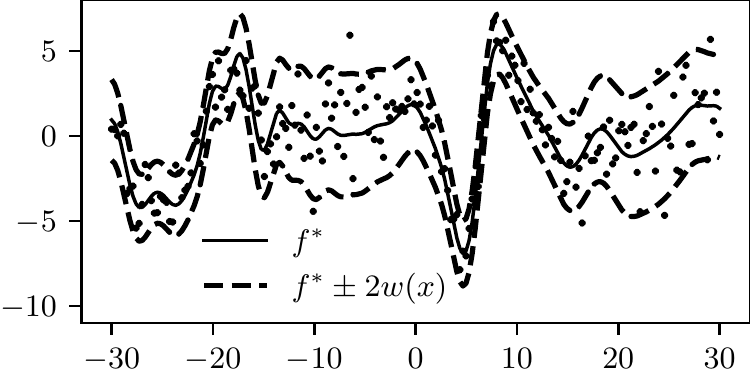}
\caption{ }
\end{subfigure}
\begin{subfigure}{0.4\textwidth}
\centering
\includegraphics[width=\linewidth]{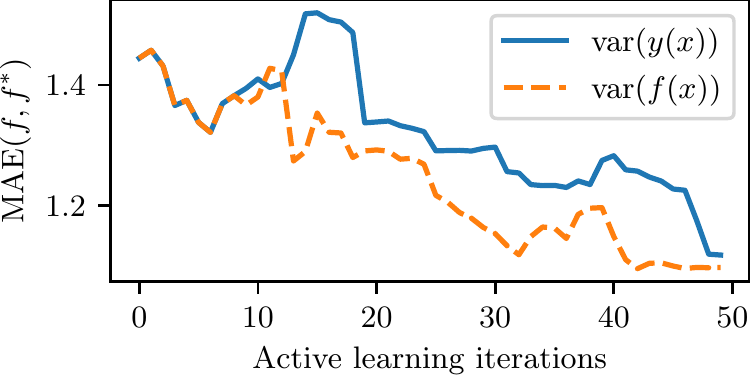}
\caption{ }
\end{subfigure}
\begin{subfigure}{0.4\textwidth}
\centering
\includegraphics[width=\linewidth]{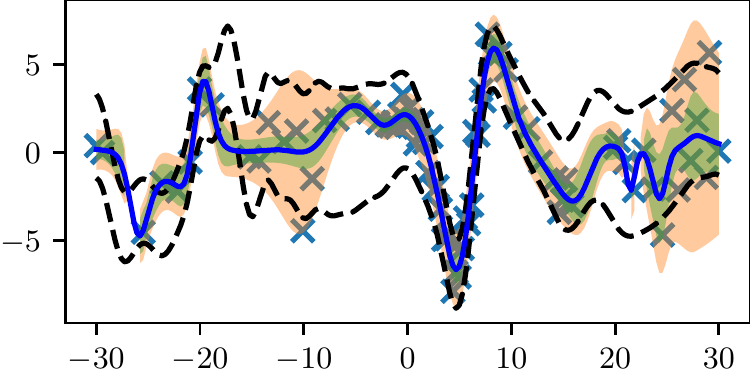}
\caption{ }
\end{subfigure}
\begin{subfigure}[b]{0.4\textwidth}
\centering
\includegraphics[width=\linewidth]{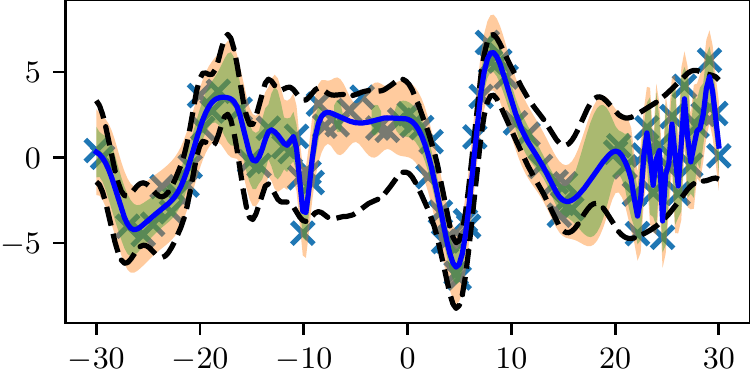}
\caption{ }
\end{subfigure}
\caption{
Fit after sampling 50 points from SYNTH-1D with active learning using (c) overall uncertainty; (d) epistemic uncertainty. We show in (b) that Mean Squared Error (MAE) between predicted function $f$ and ground truth $f^*$ is improving faster in (d) with epistemic uncertainty as compared to (c) overall uncertainty. Initial training was done on 30 points. While using epistemic uncertainty, we can capture better points that help GP learn a better fit. Black dots in (a) are data points. Green and orange regions show epistemic and overall uncertainty, respectively. 
% \nb{try density plots of samples}.
% \nb{show the f and f* function and how using epistemic uncertainty as the acquisition, we get closer to f*. We should also perhaps plot the true functions for l, w, ..also can we in addition to the qualitative argument also quantitatively show the epistemic acquisition is better..in the future/related work, perhaps we should mention mutual information?!}
}
\label{fig:ARDGM}
\end{figure}

\paragraph{Active learning}
In \cref{sec:AL}, we argued that $\var(f(\vx))$ is a better measure of uncertainty compared
to $\var(y(\vx))$, when performing active learning.
We illustrate this point on a 1d synthetic example
in \cref{fig:ARDGM}. We first train the model on 30 initial train points and then select 50 points with active learning with the following acquisition functions: c) $\var(y(\vx))$ and d) $\var(f(\vx))$. Note that we do not retrain the hyperparameters of GP during active learning to fasten the process. We empirically (b) and visually (c, d) show that points chosen by $\var(f(\vx))$ are closer to the real function.

\eat{
The following points are important for a stable optimization and reasonable results (comment: need to explain these points in a reasonable flow):
* Encourage high lengthscale and low variance with appropriate priors: Gamma(10, 1) and Gamma(0.5, 1)
* Reparametrization for non-centric learning
* Fix noise because not learning the real data
* Initialize parameters values from the respective priors.
}

\section{Related work}

Multiple approaches to create non-stationary kernels include
the multivariate generalization of the Gibbs kernel
in \cite{Paciorek2006},
sparse spectral kernels \cite{Remes2017,Tompkins2020},
the post-processing method of \cite{Risser2020},
deep kernel learning~\cite{Wilson2016aistats},
and deep GPs~\cite{Jakkala2021}.
In this paper, we adopt the Gibbs kernel approach, where the kernel parameters are themselves generated by latent GPs,
as used in prior work \cite{heinonen2016non}.
The main difference is in the model fitting procedure.
They optimize the latent hyper-parameters 
$\vphi$ by grid search,
and the compute MAP estimates of
$\tilde{\vl}=[\tilde{\ell}(\vx_n)]_{n=1}^N$,
$\tilde{\vsigma}=[\tilde{\sigma}(\vx_n)]_{n=1}^N$,
and
$\tilde{\vomega}=[\tilde{\omega}(\vx_n)]_{n=1}^N$
using gradient descent.
Thus they are optimizing a total of $3N + 9$ parameters,
namely  the outputs of the three latent hyper-functions,
and the latent hyper-parameters.
By contrast,  we optimize $2M(D+1) + 9$ parameters,
namely 
the inducing inputs to the three latent hyper-functions,
and the latent hyper-parameters.
For the problems of interest to us (spatio-temporal modeling),
$D$ is often low dimensional, so $2M(D+1) \ll 3N$,
which makes our approach faster and less prone to overfitting.
An additional advantage of our approach is that
it is easy to make predictions on new data
since we can compute the kernel $\kernelfn$ for any pair of inputs.
By contrast, in \cite{heinonen2016non}, they have to use a heuristic to extrapolate the latent hyper-functions to the test inputs,
before using them to compute the kernel values at those locations.

%\section{Discussion and Future Work}

\bibliographystyle{unsrt}
\bibliography{main}

\newpage
\appendix
\section{Model fit on various datasets}
\label{sec:appendix}

\begin{figure}[h]
\centering
\begin{subfigure}{\textwidth}
\centering
\includegraphics[width=0.9\linewidth]{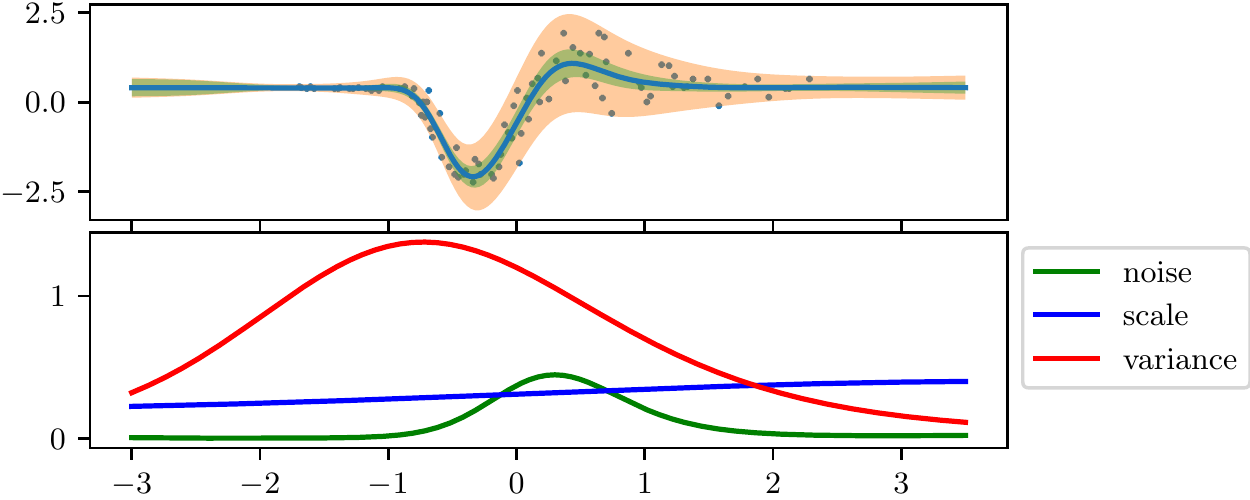}
\caption{ }
\end{subfigure}
\begin{subfigure}{\textwidth}
\centering
\includegraphics[width=0.9\linewidth]{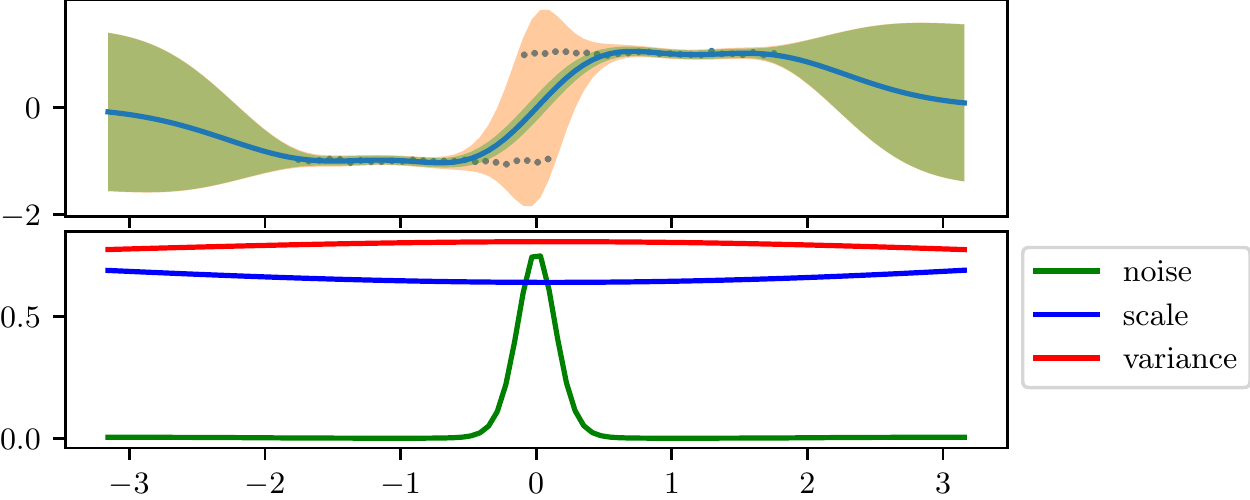}
\caption{ }
\end{subfigure}
\begin{subfigure}{\textwidth}
\centering
\includegraphics[width=0.9\linewidth]{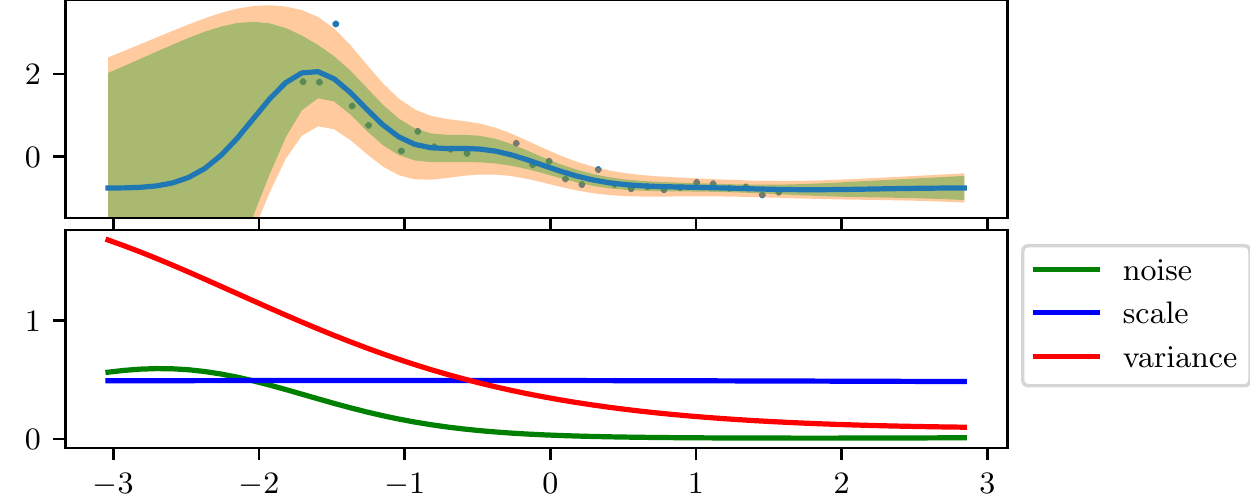}
\caption{ }
\end{subfigure}
\caption{Fitted $(\ell, \sigma, \omega)$-GP to a) motorcycle helmet data; b) step data and c) Olympic 100-m race data.
Top: predicted distribution.
Bottom: learned latent GPs for the 3 hyper-parameters: noise ($\omega$), scale ($\ell$), variance ($\sigma$). The green region is 95\% epistemic variance $\var(f(\vx))$ and orange region is overall 95\% confidence.
% \nb{we should include the mathematical symbols used in the text to make it easier for the audience.}
    }
    \label{fig:mcycle}
\end{figure}

\end{document}